\documentclass[runningheads]{llncs}
\usepackage{graphicx}
\usepackage{comment}
\usepackage{amsmath,amssymb} 
\usepackage{color}

\usepackage[colorlinks]{hyperref}
\hypersetup{final}

\newcommand{\bftab}{\fontseries{b}\selectfont}

\usepackage{soul}
\usepackage{booktabs}
\usepackage{wrapfig}
\usepackage{subcaption}
\newcommand\blfootnote[1]{%
  \begingroup
  \renewcommand\thefootnote{}\footnote{#1}%
  \addtocounter{footnote}{-1}%
  \endgroup
}

\usepackage{enumitem} 
\usepackage{verbatim}
\usepackage{mathtools}
\mathtoolsset{centercolon}

\begin{document}

\pagestyle{headings}
\mainmatter

\title{PointTriNet: Learned Triangulation of\\3D Point Sets} 

\titlerunning{PointTriNet: Learned Triangulation of 3D Point Sets}
\author{Nicholas Sharp\inst{1} \and
Maks Ovsjanikov\inst{2}}
\authorrunning{N. Sharp \& M. Ovsjanikov}
\institute{Carnegie Mellon University\and
LIX, École Polytechnique, IP Paris}

\maketitle


\newcommand{\LossChamferForward}{\mathcal{L}_{\overrightarrow{\text{C}}}}
\newcommand{\LossChamferReverse}{\mathcal{L}_{\overleftarrow{\text{C}}}}
\newcommand{\LossOverlap}{\mathcal{L}_{\text{O}}}
\newcommand{\WeightOverlap}{\lambda_{\text{O}}}
\newcommand{\LossWatertight}{\mathcal{L}_{\text{W}}}
\newcommand{\WeightWatertight}{\lambda_{\text{W}}}
\newcommand{\LossMatch}{\mathcal{L}_{\text{M}}}


\begin{abstract}
This work considers a new task in geometric deep learning: generating a triangulation among a set of points in 3D space. We present PointTriNet, a differentiable and scalable approach enabling point set triangulation as a layer in 3D learning pipelines. The method iteratively applies two neural networks: a classification network predicts whether a candidate triangle should appear in the triangulation, while a proposal network suggests additional candidates. Both networks are structured as PointNets over nearby points and triangles, using a novel triangle-relative input encoding. Since these learning problems operate on local geometric data, our method is efficient and scalable, and generalizes to unseen shape categories. Our networks are trained in an unsupervised manner from a collection of shapes represented as point clouds. We demonstrate the effectiveness of this approach for classical meshing tasks, robustness to outliers, and as a component in end-to-end learning systems.

\end{abstract}

\blfootnote{Code is available at 
\href{http://github.com/nmwsharp/learned-triangulation}{github.com/nmwsharp/learned-triangulation}
}


\section{Introduction}

Generating surface meshes is a fundamental problem in visual computing; meshes are essential for tasks ranging from visualization to simulation and shape analysis.
In this work, we specifically consider generating surfaces via \emph{point set triangulation}, where a fixed vertex set is given as input and a collection of triangles among those points is returned as output. 

Point set triangulation has been widely studied as a classical problem in computational geometry, typically in the context of surface reconstruction from sampled point clouds such as 3D range scan data \cite{khatamian2016survey,shewchuk2016delaunay,berger2017survey}.
The key property of point set triangulation is that rather than allowing \emph{any} mesh as output, the output is constrained to only those meshes which have the input points as their vertices.
When reconstructing surfaces from sensor data this constraint is artificial, and makes the problem needlessly difficult: there simply may not exist any desirable mesh which has the input points as its vertex set.
For this reason most practical methods for surface reconstruction have pivoted to volumetric techniques like Poisson reconstruction \cite{kazhdan2006poisson,kazhdan2013screened} or learned implicit functions, and extract a mesh only as a post-process \cite{park2019deepsdf,chen2019learning}.

However, recent advances in point-based based geometric learning motivate a new setting for point set triangulation, where the points are generated not as the output of a sensor, but rather as the output of a learned procedure \cite{achlioptas2017learning,rakotosaona2019pointcleannet}.
If such point sets could be easily triangulated, then any learning-based method capable of generating points could be immediately augmented to generate a mesh.
Using point-based representations avoids the implicit smoothing and computational cost that come from working in a volumetric grid, and a differentiable meshing block enables end-to-end training to optimize for the final surface mesh.

Unfortunately, classical techniques from computational geometry are not differentiable, and recent learning-based 3D reconstruction methods are typically either restricted to particular shape classes \cite{chen2019learning} or too computationally expensive to use as a component of a larger procedure \cite{dai2019scan2mesh}.  The focus of our work is to overcome these challenges, and design the first point set triangulation scheme suitable for geometric learning.  Several aspects make this problem difficult: first the method must be differentiable, yet the choice of triangles is  discrete---we will need to generalize the output to a distribution over possible triangles.  The second major challenge is scalability; exhaustive exploration is prohibitively expensive even for moderately-sized point sets. Finally, the method should be generalizable, not tied to specific shape categories.

Our main observation is that while triangulation is a global problem, it is driven largely by local considerations. 
For example, the local circumscribing-circle test in 2D is sufficient to identify faces in a Delaunay triangulation. 
Based on this intuition, we propose a novel neural network that predicts the probability that a single triangle should appear in a triangulation, and a second, similarly-structured network to suggest neighboring triangles as new candidates.
Iteratively applying these two networks generates a coherent triangle mesh.

Our approach yields differentiable triangle probabilities; gradients can be evaluated via ordinary backpropagation. The use of a proposal network allows our method to efficiently identify candidate triangles, and thus scale to very large inputs. Moreover, since our problems are local and geometric in nature, our method can easily be applied to surface patches across arbitrary shape classes. We validate these properties through extensive experiments, demonstrating that our method is competitive with classical approaches. Finally, we highlight new applications in geometric learning made possible by our differentiable approach.

\section{Related work}
\label{sec:related}

Surface reconstruction and meshing are among the oldest and most
researched areas of computational geometry, computer vision and
related fields---their full overview is beyond the scope of this
paper. Below we review methods most closely related to ours and refer
the interested readers to recent surveys and monographs
\cite{khatamian2016survey,shewchuk2016delaunay,berger2017survey} for a
more in-depth discussion.

A classical approach for surface reconstruction is to estimate
a volumetric representation, e.g. a signed distance function,
and then extract a mesh~\cite{hoppe1992surface,curless1996volumetric}. Both of these steps have been extended to improve robustness and accuracy, including
\cite{treece1999regularised,kazhdan2006poisson,mullen2010signing}
among many others (see also related techniques in
\cite{newman2006survey}). These approaches work well on  densely sampled surfaces, but both require normal
orientation, which is notoriously difficult to estimate in practice, and do
not preserve the input point samples, often leading to loss of detail.

Other approaches, based on Delaunay triangulations
\cite{boissonnat1984geometric,kolluri2004spectral,boissonnat2005provably},
alpha shapes \cite{edelsbrunner1994three,bernardini1999ball} or
Voronoi diagrams \cite{amenta1998new,amenta2001power} often come with
strong theoretical guarantees and perform well for densely sampled
surfaces that satisfy certain conditions (see
\cite{dey2006curve} for an overview of classical
techniques). Several of these approaches,
e.g. \cite{bernardini1999ball,amenta1999surface} also preserve the
input point set, but, as we show below can struggle in the presence of
poorly sampled data. Interestingly, there are several NP hardness
results for surface reconstruction
\cite{barequet1998triangulating,biedl2011reconstructing}, which both
explain the prevalence of heuristics, and point
towards the use of \emph{data-driven} priors. More fundamentally,
classical approaches are not differentiable by nature and
thus do not allow end-to-end training or backpropagation of the mesh
with respect to input positions. 

More closely related to ours are recent learning-based methods that
aim to exploit the growing collections of geometric objects, which often
come with known mesh structure. Similarly to the classical methods,
most approaches in this area are based on volumetric
representation. This includes, for example, a fully differentiable
variant of the marching cubes \cite{liao2018deep} for
computing a mesh via voxel grid occupancy
prediction \cite{gkioxari2019mesh,mescheder2019occupancy}, and recent generative models for implicit surface prediction
\cite{chen2019learning,genova2019learning}. Unlike these works, our
focus is on directly meshing an input point set. This
allows our method to scale better with input complexity, avoid
over-smoothing inherent in voxel-based techniques, and 
differentiate with respect to point positions.

Other surface generation methods include template-based techniques which fit a fixed mesh to the input \cite{litany2018deformable,kanazawa2018learning,lin2019photometric}, or deform a simple template while updating its connectivity \cite{wang2018pixel2mesh,pan2019deep}.
More recent representations include fitting parameterized surface patches to points \cite{groueix2018papier,williams2019deep}, or decomposing space in to convex sets \cite{deng2020cvxnet,chen2020bsp}.
Most importantly, none of the methods directly triangulate an arbitrary input point set.
More generally, many of these schemes are limited to specific shape categories or topologies, or are too prohibitively expensive to use as as a building-block in a larger system.

Perhaps most closely related to ours is the recent Scan2Mesh technique \cite{dai2019scan2mesh}, which uses a graph-based formulation to generate triangles in a mesh.
Though both generate meshes, there are many differences between this approach and ours, including that
Scan2Mesh is applied to a volumetric representation, as opposed to the point-based representation used here, and Scan2Mesh does not attempt to generate watertight meshes.
Several of the ideas introduced in this work could perhaps be combined with Scan2Mesh for further benefit, for instance using our proposal network (Section~\ref{sec:proposal_network}) to avoid quadratic complexity.

Finally, we note that our \emph{proposal} and \emph{classifier} architecture is inspired by classical approaches for object detection in images~\cite{girshick2014rich,ren2015faster}.

\begin{figure}
\begin{center}
\includegraphics{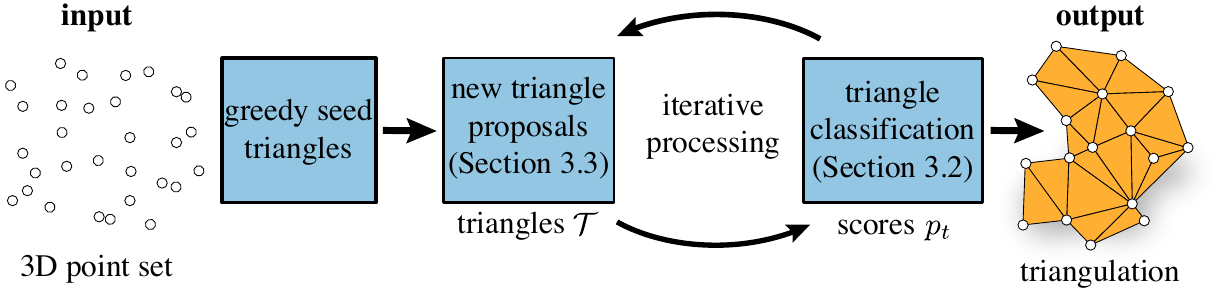}
\end{center}
\caption{An overview of our PointTriNet pipeline, which generates a triangulation by alternating between proposing new triangles and classifying them, each with a neural network. The classifier network identifies triangles which should appear in the triangulation, while the proposal network generates new candidates.}
\label{fig:method_diagram}
\end{figure}

\section{Method}
\label{sec:method}

Our method, called \emph{PointTriNet}, has two essential components, both of which are learned.
The first is a \emph{classification} network (Section~\ref{sec:class_network}), which takes a candidate triangle as input, and outputs a score, which we interpret as the probability of the triangle appearing to the triangulation.
The second is a \emph{proposal} network (Section~\ref{sec:proposal_network}), which suggests likely neighbor triangles for an existing triangle.
PointTriNet alternates between these networks, classifying candidate triangles and proposing new candidates, iteratively generating the output mesh.

\subsection{Triangle-relative coordinates}
\label{sec:encode}

Rather than treating mesh generation as a global learning problem, we pose it as a local problem of predicting a single triangle, and make many such predictions.
Both our classification and proposal subproblems are defined with respect to a particular query triangle: either classifying the triangle as a member of the mesh, or proposing candidates adjacent to the triangle.
These problems depend on both the triangle's geometry and nearby neighborhood, but it is not immediately obvious how to encode both of these quantities for a neural network while preserving the expected invariants.
Our solution is to design an encoding of points in the neighborhood of a triangle, \emph{relative} to the triangle's geometry.

For any point $p$ in the neighborhood of a triangle $t$, our encoding is given by Cartesian coordinates $x',y',z'$ in a frame aligned with the first edge and normal of $t$, as well as the point's barycentric coordinates $u,v,w$ after projecting in to the plane of the triangle.  
This results in the following encoding function, which outputs a $6$-dimensional vector for $p$ relative to $t$
\begin{equation}
\texttt{encode}\texttt{(t, p)} \to [x', y', z', u, v, w].
\end{equation}
Using both Cartesian and barycentric coordinates is essential to simultaneously encode \emph{both} the shape of the triangle and the location of a neighbor point; either individually would not suffice.
This encoding is invariant to scale and rigid transformations, but not yet invariant to the permutation of the triangle's vertices.
Per-triangle permutation invariance can be easily achieved by averaging results for all 6 possible permutations, though we find this to be unnecessary in practice.
Finally, to encode a nearby neighboring triangle using this scheme, we encode each of the neighbor's vertices, then take a max and min along these encoded values, yielding 12 coordinates.
This encoding is easy to evaluate, provides rigid- and (if desired) permutation- invariance, and is well-suited for use in point-based learning architectures.

\subsection{Classifier network}
\label{sec:class_network}

\begin{figure}[t]
\centering
\includegraphics{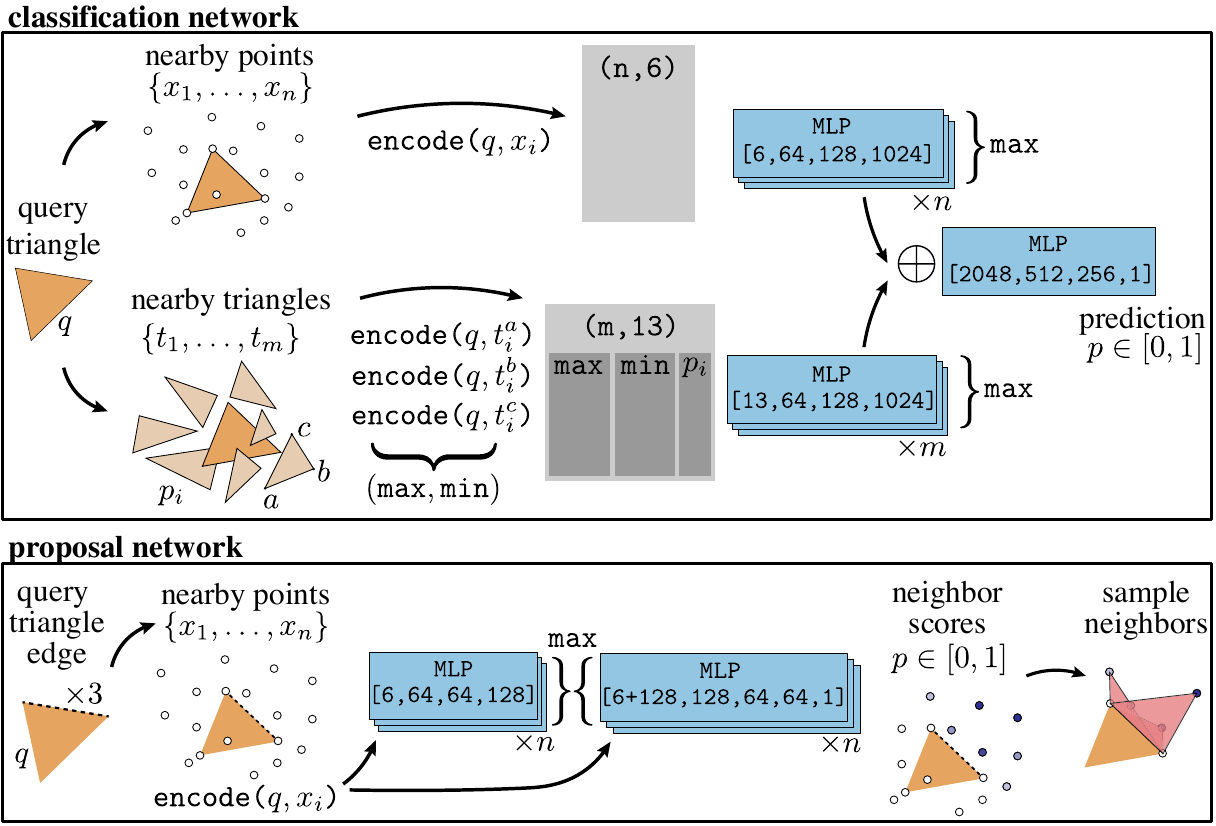}
\caption{
The core of our technique is a network which classifies whether a given triangle belongs in a triangulation, as a function of nearby points and already-classified candidates.
All coordinates are encoded relative to the query triangle. A second, similarly-structured network proposes new triangle candidates for subsequent classification. 
\label{fig:network_diagram}
}
\end{figure}

The primary tool in our method is a network which classifies whether a single query triangle belongs in the output triangulation.
This classification is a function of the nearby points, as well as nearby triangles and their previous classification scores, as shown in Figure~\ref{fig:network_diagram}.

More precisely, for a given query triangle we gather the $n$ nearest points and $m$ nearest triangles, as measured from triangle barycenters (we use $n=m=64$ throughout).
These neighboring points and triangles are then encoded relative to the query triangle as described in Section~\ref{sec:encode}, and for the nearby triangles we additionally concatenate previous classification scores for a total $12+1=13$ coordinates per triangle.
The result is a set of nearby encoded points $\mathcal{N}_\textrm{point} = \{\mathbb{R}^{6} \}_{n}$ and the set of nearby encoded triangles $\mathcal{N}_\textrm{tri} = \{\mathbb{R}^{13} \}_{m}$.
We then learn a function
\begin{equation}
    f : \mathcal{N}_\textrm{point}, \mathcal{N}_\textrm{tri} \to [0,1]
\end{equation}
which we interpret as the probability that the query triangle appears in the triangulation.
We model this function as a PointNet~\cite{qi2017pointnet}, using the multi-layer perceptron (MLP) architecture shown in Figure~\ref{fig:network_diagram}.
Here, our careful problem formulation and input encoding enables the use of a small, ordinary PointNet without spatial hierarchies or learned rotations---the inputs are already localized to a small neighborhood and encoded in a rigid-invariant manner.

\subsection{Proposal network}
\label{sec:proposal_network}

The classifier network is effective at identifying good triangles, but requires a set of candidates as input.
Naively enumerating all $O(n^3)$ possibilities does not scale, and simple heuristics miss important triangles.

Our approach identifies candidates via second \emph{proposal} network.
For each edge of a candidate triangle, the proposal network suggests new neighboring triangles across that edge.
These proposals are learned by predicting a scalar function $\mathcal{V} \to [0,1]$ on the nearby vertices.
Formally, for an edge $ij$ in triangle $ijk$, we learn the probability that a nearby vertex $l$ forms a neighboring triangle $ijl$.
These probabilities are then sampled to generate new candidates, which are assigned an initial probability as the product of the probability for the triangle that generated them and their learned sampling probability. 

The proposal network is again a PointNet over the local neighborhood, though it takes only points as input, and outputs a per-point result.
We use the same triangle-relative point encoding (Section~\ref{sec:encode}), but here the ordering of query triangle vertices $ijk$ serves as a feature---the network always predicts neighbors across edge $ij$.
Applying the network to the three cyclic permutations of $ijk$ yields three sets of predictions, potential neighbors across each edge.

\subsection{Iterative prediction}

\begin{figure}
\centering
\includegraphics{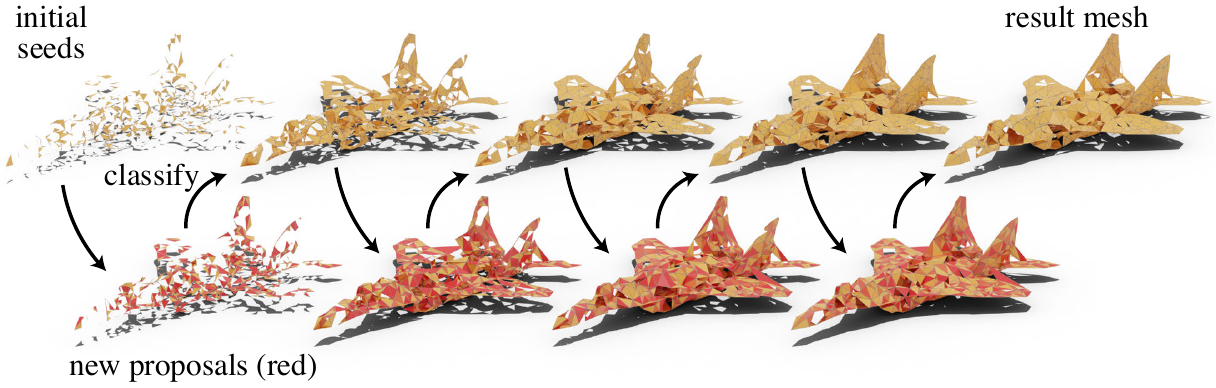}
\caption{Our approach iteratively constructs a triangulation, alternating between classifying triangles and proposing new candidates. Classified triangles are drawn in orange, and new proposals are drawn in red, for both only triangles with $p > 0.5$ are shown. The input vertex set is 1k sampled points, not shown here.
\label{fig:iterative_meshing}
}
\end{figure}

To triangulate a point set, we alternate between proposing new candidates and classifying them.
At each iteration, each candidate triangle $t$ has probability $p_t \in [0,1]$, the probabilities from each round are features in the next, akin to a recurrent neural network.
We initialize with greedy seed triangles among the nearest neighbors of each vertex, then for each iteration we first classify candidates, then generate new candidates, retaining only the highest-probability candidates.
If multiple instances of the same triangle arise, we discard lower probability instances.
In our experiments, we iterate $5$ times, sample 4 neighbors across each edge per iteration, and retain $12 |V|$ candidates.
If desired, the output can be post-processed to fill small holes (Figure~\ref{fig:generalization_gallery},\ref{fig:offset}, \emph{right}).

\subsection{Losses and training}
\label{subsec:losses}
Our method outputs a probabilistic surface, where each triangle has an associated probability.
This is a necessary generalization: the choice of triangles is a discrete set, and thus does not have traditional smooth derivatives.
We thus make use of probabilistic losses detailed below, which are essentially the expected value of common geometric losses, like expected Chamfer distance.

For training, we assume a known, ground-truth surface $\mathcal{S}$, and the probabilistic surface $\mathcal{T}$ with probabilities $p:\mathcal{T} \to [0,1]$ predicted by our algorithm.
Note that we do not assume to be given a ground truth triangulation of $\mathcal{S}$, only some representation of the underlying surface sufficient to measure distance.
In the losses below we assume $\mathcal{S}$ is represented by a sampled point cloud, but other representations like an implicit function could easily be substituted as training data.
Notice that our method is thus unsupervised, in the sense that it learns to generate triangulations given un-annotated point clouds as input---it is not trained to match existing triangulations.

\paragraph{Expected Chamfer distance (forward)}
The distance from the ground truth surface to our predicted surface is measured as
\begin{align}
\LossChamferForward
&:= E[\int_{x \in \mathcal{S}} \min_{y \in \mathcal{T}} d(x,y)]
= \int_{x \in \mathcal{S}} E[\min_{y \in \mathcal{T}} d(x,y)] \nonumber \\
&= \int_{x \in \mathcal{S}} \int_{y \in \mathcal{T}} \gamma(x,y) d(x,y)
\end{align}
where $\gamma(x,y)$ denotes the probability that $y$ is the closest point in $\mathcal{T}$ to $x$.
Discretely, we evaluate this by sampling points $x$ on $\mathcal{S}$, then for each point sorting all triangles in $\mathcal{T}$ by distance from $x$. 
A cumulative product of probabilities for these sorted triangles gives the probability that each triangle is the closest to $x$, and taking an expectation of distance under these probabilities yields the desired expected distance from the sample to our surface. 
In practice we truncate to the $k=32$ nearest triangles.

\paragraph{Expected Chamfer distance (reverse)}
The distance from our predicted surface to the ground truth surface is measured as
\begin{align}
\LossChamferReverse
&:= E[\int_{y \in \mathcal{T}} \min_{x \in \mathcal{S}} d(x,y)]
= \int_{y \in \mathcal{T}} p(y) \min_{x \in \mathcal{S}} d(x,y)
\end{align}
Discretely, we evaluate this by sampling points $y$ on $\mathcal{T}$, then for each point we measure the distance to $\mathcal{S}$ and sum, weighting by the probability $p(y)$.
Similar losses have appeared in 3D reconstruction, e.g.{} \cite{liao2018deep}.

\paragraph{Overlap kernel loss}
To discourage triangles from overlapping in space, we define a spatial kernel around each triangle
\begin{equation}
g_t(x) = p_t \max\left(0, 1 - \dfrac{d_n(x)}{d_e(x)}\right)
\end{equation}
where  $x$ is an arbitrary point in space, $p_t$ is the probability of triangle $t$, $d_n$ is the distance in the normal direction from the triangle, and $d_e$ is the signed perpendicular distance from the triangle's edge (see supplement for details).
In a good triangulation, for any point $x$ on the surface there will be exactly one triangle $t$ for which $g_t(x) \approx 1$ while $g_{t'}(x) \approx 0$ for all other triangles $t'$.
This is modeled by the loss
\begin{equation}
\LossOverlap := \int_{x \in \mathcal{T}} \left(-1 +  \sum_{t \in \mathcal{T}} g_t(x)\right)^2 + \left(-1 +  \max_{t \in \mathcal{T}} g_t(x)\right)^2
\end{equation}
which is minimized when exactly one kernel contributes a value of $1$ at $x$. 
Discretely, we sample points $x$ on the surface of the generated triangulation $\mathcal{T}$ to evaluate the loss.
One could instead sample $x$ on the ground truth surface, but using the generated triangulation $\mathcal{T}$ makes the loss a regularizer, applicable in generative settings.

\paragraph{Watertight loss}

To encourage a watertight mesh, we explicitly maximize the probability that each edge in the triangulation is watertight, via
\begin{equation}
  \LossWatertight := \sum_{ij \in \mathcal{E}} p_{ij} (1 -  p_{ij}^{\textrm{water}})
\end{equation}
where $\mathcal{E}$ is the set of edges, $p_{ij}$ denotes the probability that $ij$ appears in the triangulation, and $p_{ij}^{\textrm{water}}$ denotes the probability that $ij$ is watertight.
An expression for evaluating this loss from triangle probabilities is given in the supplementary material.
Notice that this loss does not directly penalize vertex-manifoldness---we observe that in practice almost all watertight configurations are also vertex-manifold, and thus simply encourage watertightness.

\paragraph{Network training}

The full loss for training the classification network is given by
\begin{equation}
\mathcal{L} := \LossChamferForward + \LossChamferReverse + \WeightOverlap \LossOverlap + \WeightWatertight \LossWatertight,
\end{equation}
where we use $\WeightOverlap = 0.01$ and $\WeightWatertight = 1$ for all experiments.
All losses are normalized by the surface area or number of elements as appropriate, for scale invariance.
During training, we backpropagate gradients only through the last iteration of classification.
This strategy proves effective here because the network must be able to classify the most useful triangles regardless of the quality of the current candidate set.

We train the proposal network simultaneously with the classifier by encouraging its suggestions to match the classification scores.
Intuitively, this means that the network attempts to propose triangles which will receive a high classification score on the next iteration.
During training, we perform one last round of proposal to generate a suggestion set ${P}$ with probabilities $u_t$.
Rather than merging these triangles into the candidate set ${T}$ as usual, we instead evaluate the loss
\begin{equation}
\LossMatch := \frac{1}{|{P} \cap {T}|} \sum_{t \in {P} \cap {T}} (u_t - p_t)^2
\end{equation}
and backpropagate through this loss only with respect to the proposal scores $u_t$.
During training we also augment the proposal network to predict a random neighbor with 25\% probability, to encourage diversity in the proposal set.

\section{Experiments}
\label{sec:experiments}

\paragraph{Architecture and training}
The layer sizes of our networks are given in Figure~{\ref{fig:network_diagram}}; we use ReLU activation functions throughout~\cite{nair2010rectified}, except for a final single sigmoid activation to generate predictions. 
We use dropout with $p=0.5$ on only the hidden layers of the prediction MLPs~\cite{srivastava2014dropout}.
Neither batch nor layer normalization are used, as they were observed to negatively impact both result quality and the rate of convergence.

We train using the ADAM optimizer with a constant learning rate of 0.0001, and batch size of $8$.
Our networks are trained for 3 epochs over 20k training samples, which amounts to $\approx\! 5$ hours on a single RTX 2070 GPU.

\paragraph{Datasets}
We validate our approach on the ShapeNetCore V2 dataset~\cite{shapenet2015}, which consists of about $50,\!000$ 3D models.
We use the dataset-recommended 80\%/20\% train/test split, and further reserve 20\% of the training models as a validation set.  
Our method is geometric in nature, so unlike semantic networks which operate per-category, we train and test simultaneously on all of ShapeNet, and do not need to subdivide by category.

\begin{figure}
\begin{center}
\includegraphics[width=\linewidth]{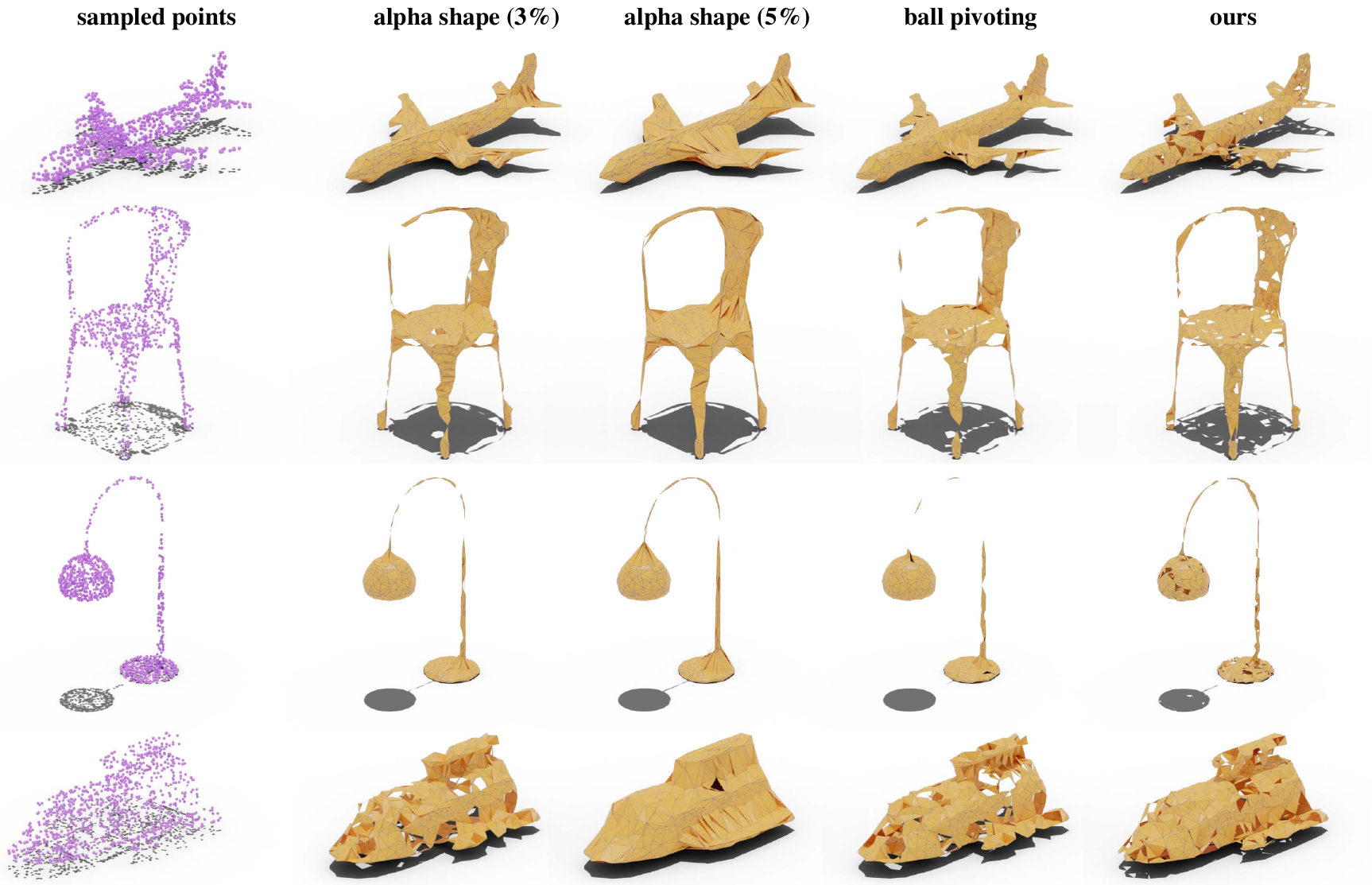}
\end{center}
\caption{A selection of outputs from PointTriNet and baselines on ShapeNet. The input is 1k points sampled on the surface of the shape, and the output is a mesh using those points as vertices. Constructing point set triangulations which are both accurate and watertight is still an open problem, but our method shows that a differentiable, learning-based approach can yield results competitive with classical computational geometry schemes.}
\label{fig:shapenet_comparison_gallery}
\end{figure}

\begin{figure}
\begin{center}
\includegraphics[width=\linewidth]{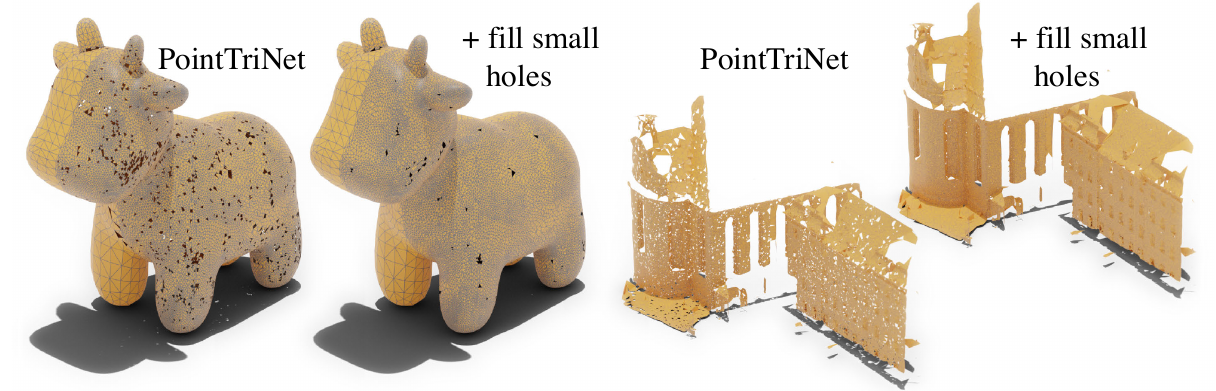}
\caption{Our method is data-driven, but its geometric nature leads to excellent generalization. \emph{Left:} reconstructing  a synthetic model with vertices of varying regularity and density. \emph{Right:} reconstructing a 3D range scan of a cathedral (from~\cite{hackel2017semantic3d}). Both use our networks trained on uniformly-sampled ShapeNet in Section~\ref{sec:experiments}.
\label{fig:generalization_gallery}
}
\end{center}
\end{figure}

\begin{table}[b]
\centering
\caption{When evaluating on sampled ShapeNet, PointTriNet outperforms classical methods in geometric accuracy, though it has moderately less watertight connectivity.  \label{tab:results_shapenet}}
\begin{tabular}{p{2.5cm}p{1.5cm}p{1.5cm}p{1.5cm}}
\toprule
{} &  Chamfer &  watertight &    manifold \\
\midrule
ours        & \bftab 0.7417 &   77.0\% &   97.4\% \\
ballpivot   &   1.3440 &  \bftab 84.1\% & \bftab 100.0\% \\
alpha-3     &   1.1099 &   49.8\% &   61.3\% \\
alpha-5     &   0.9195 &   47.6\% &   53.0\% \\
\bottomrule
\end{tabular}
\end{table}

We uniformly sample 1k points on the surface of each model, which will serve as the vertex set to be triangulated, and then separately sample another 10k points from each model, which serve as the representation of the surface when evaluating loss functions.  
Note that we \emph{do not} use the mesh structure of the dataset during our training procedure.

To generate training examples in small neighborhoods, we choose a random center point on some shape, and gather the 256 nearest vertex samples (among the 1k), along with all surface samples (among the 10k) in the same radius.
We use 20k such neighborhoods as a training set, with another 5k for validation.
These neighborhoods are used only for efficient training---to evaluate our method on the test set we triangulate the full 1k sampled vertices for each model.

\paragraph{Baselines}
We compare our learned approach against two classical methods for point set triangulation: ball pivoting and $\alpha$-shapes~\cite{edelsbrunner1994three,bernardini1999ball}. 
For ball-pivoting, the ball radius is automatically guessed as the bounding box diagonal divided by the square root of the vertex count.
For $\alpha$-shapes, we report two different choices of the radius parameter $\alpha$, as 3\% and 5\% of the bounding box diagonal.
No prior learning-based methods can directly serve as a baseline, but in the supplement we adapt a variant of Scan2Mesh~\cite{dai2019scan2mesh} for further comparison.

\paragraph{Metrics}

Geometric accuracy is assessed with a bi-directional Chamfer distance
\begin{equation}
\frac{1}{|\mathcal{A}|} \int_{x \in \mathcal{A}} \min_{y \in \mathcal{B}} d(x,y) + \frac{1}{|\mathcal{B}|} \int_{y \in \mathcal{B}} \min_{x \in \mathcal{A}} d(x,y)
\end{equation}
and evaluated discretely by sampling 10k points on both meshes.
Reported Chamfer distances are scaled by $100\times$ for readability.
We measure mesh connectivity by \emph{watertightness}, the percentage of edges which have exactly two incident triangles, and (edge-) \emph{manifoldness}, the percentage of edges with one or two incident triangles---with this terminology mesh boundaries are manifold but not watertight. 
For our method, we take the triangles with $p > 0.9$ as the output.

\subsection{Results}

The most basic task for PointTriNet is to generate a mesh for a point set sampled from some underlying shape.
Table~\ref{tab:results_shapenet} shows the effectiveness of our method on the sampled ShapeNet dataset, compared to the baseline approaches.
Our method is geometrically more accurate than all of the classical schemes, though it achieves somewhat less regular connectivity than ball pivoting.
We note that it may not be possible for any algorithm to significantly improve on these metrics: an imperfect (e.g., random) vertex sampling likely does not admit any triangulation which has both perfect geometric accuracy and perfect manifoldness.
The supplement contains an ablation study, a breakdown by class, and triangle quality statistics.

\paragraph{Performance}
Triangulating a point set with PointTriNet amounts to evaluating standard MLPs, in contrast to methods which e.g.\ solve an optimization problem for each input (Section \ref{sec:related}).
In our unoptimized implementation, triangulating 1000 points takes about $1$ second on an RTX 2070 GPU.
To reduce the memory footprint for large point sets (Figure \ref{fig:generalization_gallery}), one can evaluate the MLPs independently over many small patches of the input.

\paragraph{Outliers}
An immediate benefit of a learned approach is that it can adapt to deficiencies in the input data.
We add synthetic noise to our sampled ShapeNet dataset by perturbing 25\% of the points according to a Gaussian with 2\%-bounding-box deviation, and train our method from scratch on this data (Table~\ref{tab:results_shapenet_noise}).
The classical baselines are not designed to handle noise, so our approach yields an even larger improvement in geometric accuracy in this setting.

\begin{table}
\centering
\caption{Evaluation on sampled ShapeNet, with noise added to a subset of the samples. Our method can adapt to the noise, further improving geometric accuracy.\label{tab:results_shapenet_noise}}
\begin{tabular}{p{1.5cm}p{1.5cm}p{1.5cm}p{1.5cm}}
\toprule
{} &  Chamfer &  watertight &    manifold \\
\midrule
ours      &   \bftab 1.0257 &   74.3\% &   98.0\% \\
ballpivot &   1.5418 &    \bftab 83.9\% &  \bftab 100\% \\
alpha-3    &   1.3260 &   50.8\% &   61.5\% \\
alpha-5    &   1.2309 &   47.5\% &   52.7\% \\
\bottomrule
\end{tabular}
\end{table}

\begin{figure}
\begin{center}
\includegraphics[width=\linewidth]{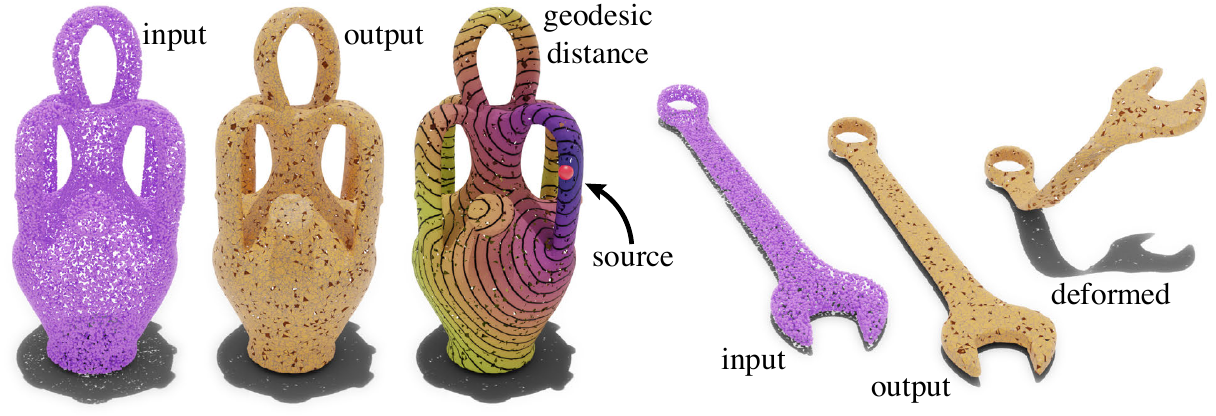}
\caption{PointTriNet directly generates a triangle mesh, opening the door to many standard geometric algorithms. \emph{Left}: geodesic distance is computed from a source point \cite{Crane:2017:HMD}. \emph{Right}: a reconstructed object is deformed via an anchor at one endpoint \cite{lipman2004differential}.
\label{fig:geometry_processing_gallery}
}
\end{center}
\end{figure}

\paragraph{Geometry processing}
A key benefit of our strategy is that PointTriNet directly generates a standard triangle mesh.
Although these meshes are not necessarily manifold (a property shared with classical methods like $\alpha$-shapes), they are still suitable for many standard algorithms in geometry processing (Figure~\ref{fig:geometry_processing_gallery}).

\begin{figure}[b]
\begin{center}
\includegraphics[width=\linewidth]{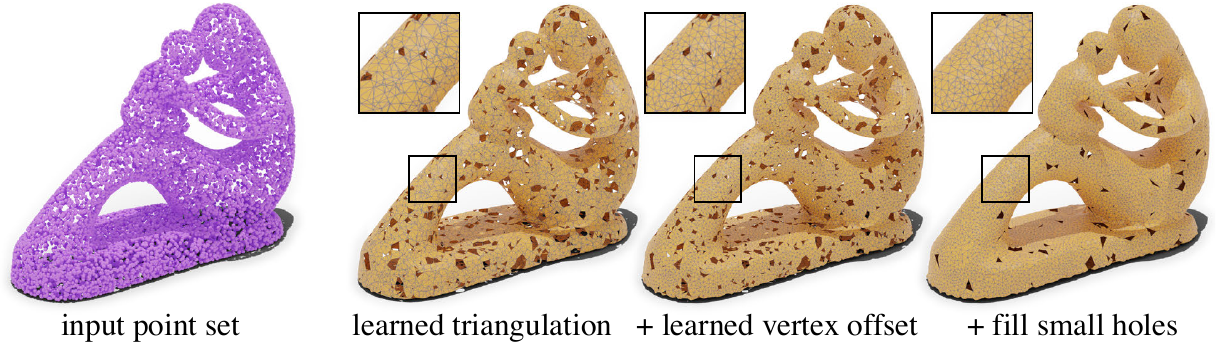}
\caption{Our method can be incorporated with other learning-based techniques. Here, we additionally learn a position offset at points to further improve output mesh quality.  \label{fig:offset}}
\end{center}
\end{figure}

\subsection{PointTriNet in geometric learning}
Our method is designed to be a new building block in geometric learning.

\paragraph{Learned vertex improvement}

We design a network which additionally learns a position update at each of the input vertices to further improve the quality of the output mesh, inspired by methods like \cite{rakotosaona2019pointcleannet}.
More precisely, we introduce an additional MLP per-triangle, structured identically to the prediction MLP except the last layer has dimension $\mathbb{R}^{3\times2}$: an offset in the triangle's tangent basis for each of its three vertices, accumulated to yield a per-vertex update in the output mesh.
This network, coupled with the base PointTriNet, is trained end to end as described above, including an additional loss term which encourages mesh quality by penalizing the deviation of the edge lengths.
Figure~\ref{fig:offset} shows how these learned offsets further improve mesh quality with a more natural distribution of vertices.

\paragraph{Generative triangulation}

\begin{figure}[t]
\begin{center}
\includegraphics[width=\linewidth]{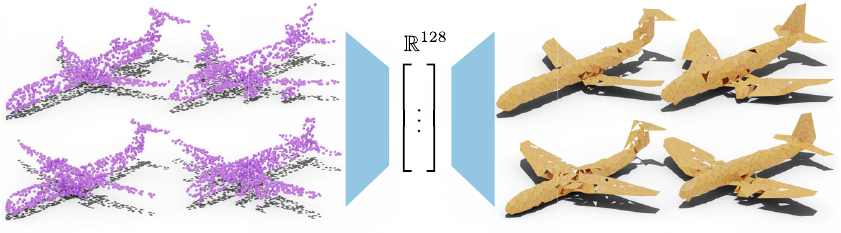}
\caption{A preliminary example of a an autoencoder, which uses our method to directly output a mesh without any intermediate volumetric representations.
Input point clouds (\emph{left}) are decoded to meshes with 512 vertices (\emph{right}).}
\label{fig:autoencoder}
\end{center}
\end{figure}

PointTriNet enables triangulation as a component in end-to-end systems---algorithms designed to output points can be augmented to generate meshes useful for downstream applications.
As a proof-of-concept, we construct a shape autoencoder which directly outputs a mesh using our technique.  
The encoder is a PointNet mapping input points to a 128 dimensional latent space, then decoding to 512 vertices with the architecture of \cite{achlioptas2017learning}; these vertices are then triangulated with PointTriNet.
We evaluate this scheme on the planes class of ShapeNet, as shown in Figure~\ref{fig:autoencoder}.
The point encoder/decoder and PointTriNet are pretrained as in \cite{achlioptas2017learning} and Table \ref{tab:results_shapenet}, respectively, and the full system is then trained end-to-end for 3 epochs with the losses in Section \ref{subsec:losses}.

This autoencoder is intentionally simplistic, and we leave a full investigation of its effectiveness to future work.
However, it serves to demonstrate that PointTriNet can indeed directly generate mesh outputs from point-based geometric learning, while nearly all past approaches instead relied on volumetric representations or template deformations.

\section{Limitations and future work}

We introduce PointTriNet, a data-driven approach to address the classical problem of triangulating a given set of points.  Our scheme uses point-based neural network architectures, and is shown to be accurate, generalizable, and scalable. 

The meshes resulting from our method are competitive in accuracy with state-of-the-art point set triangulation methods from computational geometry. Crucially, however, our approach is fully differentiable, enabling novel applications such as building an autoencoder network with a triangle mesh output.

Our meshes are still noticeably less smooth than those extracted from volumetric representations---this is an inherent challenge of the strictly-harder point set triangulation problem, as opposed to general surface reconstruction. By incorporating topological priors, hole-filling, and building upon the differentiability of the meshing network, future work could further improve mesh quality.

We believe that this new approach to a classical problem will spur further research in geometric learning, and prove to be a useful component in a wide variety of applications.

\paragraph{Acknowledgements}
The authors are grateful to Marie-Julie Rakotosaona and Keenan Crane for fruitful initial discussions, and to Angela Dai for assistance comparing with Scan2Mesh.
Parts of this work were supported by an NSF Graduate Research Fellowship, the KAUST OSR Award No. CRG-2017-3426 and the ERC Starting Grant No. 758800 (EXPROTEA).

%
%
\bibliographystyle{splncs04}
\bibliography{lc}

\newpage

\appendix

\section{Loss functions}

These loss functions are introduced in the main document. Here we give additional details and implementation notes.

\subsection{Overlap loss}

This loss is used to penalize triangles which overlap in space.
The key ingredient is a spatial kernel, the definition of which is reproduced here as
\begin{equation}
\label{eq:kernel_eq}
g_t(x) := p_t \max(0, 1 - \dfrac{d_n(x)}{d_e(x)}),
\end{equation}
\setlength{\columnsep}{1em}
\setlength{\intextsep}{0em}
\begin{wrapfigure}{r}{105pt}
   \includegraphics[width=105pt]{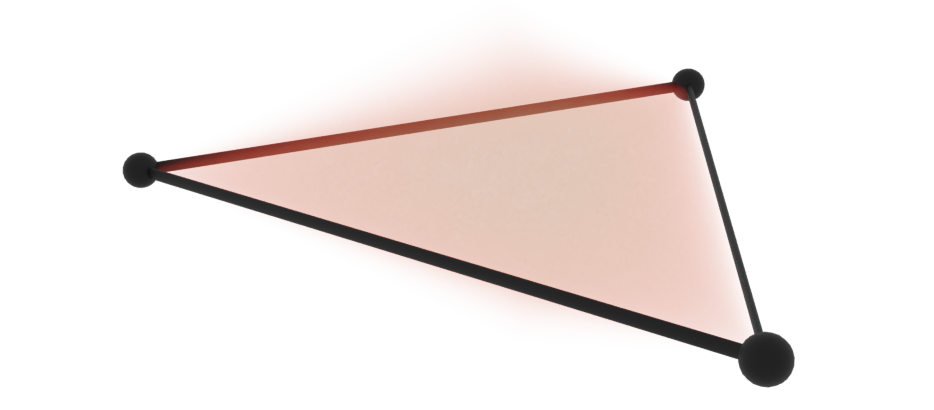}
   \caption{A volumetric rendering of the overlap kernel from Equation~\ref{eq:kernel_eq}.}
   \vspace{.4em}
\end{wrapfigure}
for any point $x \in \mathbb{R}^3$, where $p_t$ is the triangle probability, $d_n(x)$ is the distance in the normal direction from the triangle, $d_e(x)$ is the smallest signed perpendicular distance to the triangle's edges, and we let $g_t(x) := 0$ for points where $d_e(x) \leq 0$.

More precisely, suppose a triangle has vertices $(p_0, p_1, p_2)$ and unit normal $n$.
Then $g_t(x)$ could be computed, for instance, via the expressions
\begin{equation}
d_n(x) := |(x - p_0) \cdot n|,
\end{equation}
where $\cdot$ denotes the dot product, and
\begin{equation}
d_e(x) := \min
\begin{cases}
{\displaystyle\min_{i \in 0,1,2}} (x - p_i) \cdot (\frac{p_{i+1} - p_i}{||p_{i+1} - p_i||} \times n)\\
0
\end{cases}
\end{equation}
where the index in $p_{i+1}$ is taken modulo 3, and the outer $\min(\dots, 0)$ automatically ensures that $g_t(x)$ takes a value of $0$ for points which are outside of the triangle when projected in to the triangle plane.

\newpage
\subsection{Watertight loss}

In a watertight and manifold mesh, all edges have exactly two incident triangles.
Accordingly, our watertightness loss penalizes edges that have a number of incident triangles other than two (that is, just one or more than two incident triangles).
Because we work with probabilistic surfaces, the loss is the expected fraction of such non-watertight edges (reproduced here)
\begin{equation}
  \LossWatertight := \sum_{ij \in \mathcal{E}} p_{ij} (1 -  p_{ij}^{\textrm{water}})
\end{equation}
where $\mathcal{E}$ is the set of edges, $p_{ij}$ denotes the probability that $ij$ appears in the triangulation, and $p_{ij}^{\textrm{water}}$ denotes the probability that $ij$ is watertight.  
As noted in the main text, this loss only captures connectivity across edges, and thus does not explicitly discourage vertex-nonmanifold triangulations.
However, we observe that practice, there are very few configurations which are watertight but not vertex-manifold (essentially just an ``hourglass'' configuration).
Since such configurations are very particular and rare, this watertightness loss seems to be sufficient to encourage vertex-manifold triangulations in practice.  

To evaluate this quantity discretely, we decompose it as a sum over halfedges $h$, where the term \emph{halfedge} refers to a side of a triangle (in a manifold mesh, each of these sides forms \emph{half} of an edge).
Each triangle has three halfedges.
As always, we will make the approximation that all triangle probabilities are independent.

Recall that $p_t$ denotes the probability associated with triangle $t$.
For any halfedge $h$, let $\tau(h)$ be the triangle containing $h$, and $e(h)$ be the edge on which $h$ is incident.
For any edge $g$, let $\mathcal{N}_g$ be the set of halfedges incident on $g$, and we will say that these halfedges are \emph{neighbors}.
For a halfedge $h$, the probability that there is exactly one other neighbor halfedge incident on the same edge is given by
\begin{equation}
    p_{h}^{\textrm{1 other}} := 
    \sum_{\mathclap{\substack{k \in \mathcal{N}_{e(h)} \\ k \neq h}}} 
    \bigg(  p_{\tau(k)} 
    \prod_{\mathclap{\substack{j \in \mathcal{N}_{e(h)} \\ j \neq h, k}}}
    1-p_{\tau(j)}  \bigg).
\end{equation}
In this expression, the outer sum is over all halfedges $k$ which could be neighbors of $h$, and for each it computes the probability that $k$ is present and is the only neighbor of $h$.

The full loss is then computed as 
\begin{equation}
\LossWatertight \gets \frac{\sum_h p_{\tau(h)} * (1 - p_{h}^{\textrm{1 other}})}{\sum_h  p_{\tau(h)}}
\end{equation}
where we normalize the expectation as a fraction of halfedges, because the probabilistic surface does not have an obvious number of edges. 

\newpage
\section{Comparison to Scan2Mesh}

This work is the first to directly consider the triangulation of point sets via machine learning.
The most related learning-based approach is the recent Scan2Mesh (see main document for citation).
There are significant differences between the problem considered in this work and the problem considered in Scan2Mesh.
Most importantly, Scan2Mesh does not attempt to triangulate arbitrary input vertex sets, or scale beyond a few hundred elements.
Additionally, Scan2Mesh uses volumetric signed-distance data as input, and is not designed to operate on unstructured point cloud input as considered here. 
Nonetheless, in the interest of comparison, we construct a network inspired the architecture in Scan2Mesh and apply it to our task.

We will refer to this comparison approach as \emph{Scan2Mesh-like}, to distinguish it from the original work.
In the style of the two-phase triangle prediction network in Scan2Mesh, we first form the $k$-nearest-neighbor graph among input vertices with $k=16$, and apply a message-passing graph network.
This network is identical to the architecture used in Scan2Mesh, except that the inputs are simply the vertex positions, and the output is an edge probability in $[0,1]$.
We then form all possible triangles among the resulting edges, and assign each an initial probability as the product of the three edge values.
For the second phase, we construct another graph network among the dual graph of these triangles, and compute per-triangle input features as in Scan2Mesh.
This graph network predicts a new probability for each triangle, which is multiplied by the initial triangle probability to generate an output value.
Scan2Mesh is trained in part using cross-entropy losses against simplified target meshes; to apply Scan2Mesh-like to our task, we instead apply our unsupervised loss functions, and train with the dataset and methodology described in Section 4 until convergence.

We emphasize that Scan2Mesh-like has many differences from Scan2Mesh, and should not be considered an implementation of that work, merely a similar method inspired by Scan2Mesh.
Differences include:
\begin{itemize}
    \item Scan2Mesh as presented performs dense $|V| \times |V|$ edge prediction. Scan2Mesh-like predicts edges on just the $k$ nearest-neighbors, to enable scaling to vertex sets of size $|V| = 1000$ at the cost of likely missing some edges and triangles.
    \item Scan2Mesh is trained using a multi-stage, supervised loss function, while Scan2Mesh-like is trained using the probabilistic loss functions and point dataset presented in this work.
    \item The triangulation networks in Scan2Mesh access data on a volumetric grid, while Scan2Mesh-like operates solely on points.
\end{itemize}

\begin{figure}
\begin{center}
\includegraphics[width=0.65\linewidth]{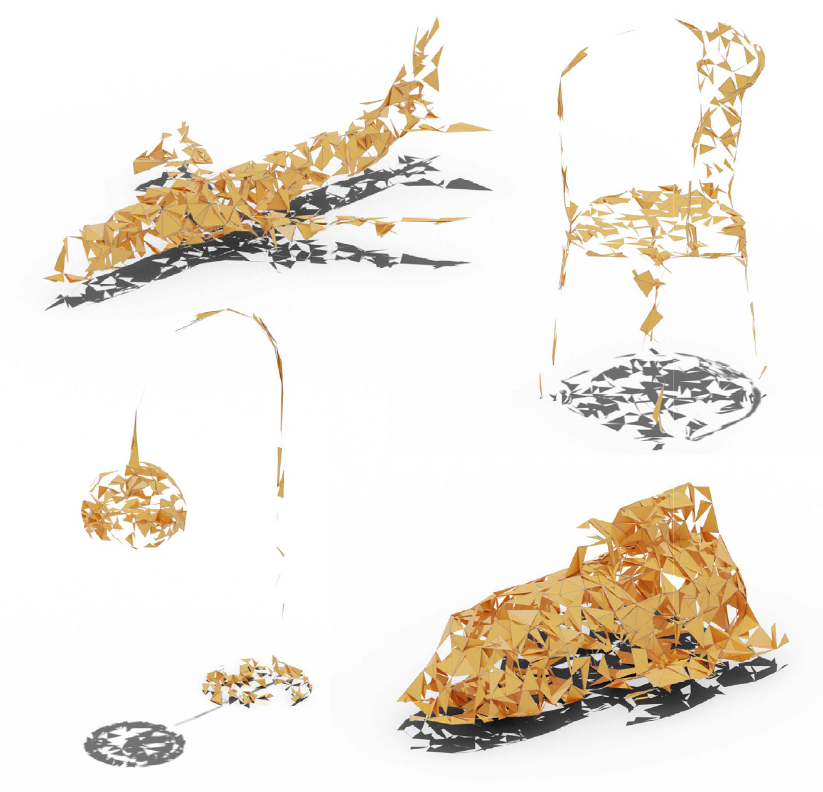}
\caption{Meshes resulting from the Scan2Mesh-like network, corresponding to Figure 4 from the main document. Attempting to generate watertight, manifold meshes yields incomplete triangulations. One cause is that in an attempt to scale Scan2Mesh to 1000s of input vertices, we consider only triangles formed among the $16$ nearest-neighbor vertices, which omits important triangles (using $k > 16$ is prohibitively expensive). In our method, the proposal network solves this issue by scalably generating good candidates.}
\label{fig:scan2mesh_gallery}
\end{center}
\end{figure}

\begin{table}
\caption{Evaluation of the Scan2Mesh-like network for the triangulation task on uniformly-sampled ShapeNet.\label{tab:results_scan2mesh}}
\centering
\begin{tabular}{lrrr}
\toprule
{} & Chamfer $\times 100$ & watertight & manifold \\
\midrule
ours &              \bftab 0.7417 &    \bftab 77.0\% &  \bftab 97.4\% \\
scan2mesh-like &               1.1887 &     23.3\% &   96.5\% \\
\bottomrule
\end{tabular}
\end{table}

Table~\ref{tab:results_scan2mesh} gives the result of evaluating the trained Scan2Mesh-like on our dataset, and Figure~\ref{fig:scan2mesh_gallery} visualizes some samples. 
The task and architecture of Scan2Mesh-like have several differences from Scan2Mesh, but this experiment serves as some basic evidence that the iterative, PointNet-based architecture presented in this work can outperform the dual-graph prediction network from Scan2Mesh.

\newpage

\section{Ablation Study}

We justify the components of our approach via an ablation study (Table~\ref{tab:results_ablation}).
The \verb|no_tris| variant omits the neighboring triangles in the classification network.
The \verb|no_proposal| variant omits the proposal network, and instead samples new neighbors heuristically, preferring close neighbors on the correct side of each edge.
The \verb|no_overlap| variant omits the overlap loss, setting $\WeightOverlap=0$.
We find that both the neighboring triangles and the proposal network are critical for high-quality connectivity in the results, removing either significantly degrades the watertightness.
The overlap term contributes a small improvement to all metrics.

\begin{table}
\centering
\caption{An ablation study over the components of our method on ShapeNet. Both the proposal network and neighboring triangles improve watertightness.\label{tab:results_ablation}}
\begin{tabular}{p{2.5cm}p{1.5cm}p{1.5cm}p{1.5cm}}
\toprule
{} &  Chamfer &  watertight &    manifold\\
\midrule
full method          &   \bftab 0.7417 &   \bftab 77.0\% &   \bftab 97.4\% \\
\verb|no_tris|       &   0.8310 &   64.6\% &   95.7\% \\
\verb|no_proposal|   &   0.7748 &   66.4\% &   97.1\% \\
\verb|no_overlap|    &   0.7489 &   76.0\% &   96.8\% \\
\bottomrule
\end{tabular}
\end{table}

\newpage
\section{Extended results}

\begin{table*}[!b]
\vspace{-1em}
\begin{subtable}[t]{0.45\textwidth}
\centering

\begin{tabular}{lrrr}
\multicolumn{4}{c}{\textbf{table}} \\
\toprule
{} & Chamfer & watertight & manifold \\
\midrule
ours      &               0.8536 &     76.3\% &   97.4\% \\
ballpivot &               1.4081 &    \bftab 82.5\% & \bftab 100.0\% \\
alpha3    &               1.1418 &     50.5\% &   61.8\% \\
alpha5    &               \bftab 0.7649 &     48.2\% &   53.3\% \\
\bottomrule
\end{tabular}
\end{subtable}\hspace{.1\textwidth}%
\begin{subtable}[t]{0.45\textwidth}
\centering

\begin{tabular}{lrrr}
\multicolumn{4}{c}{\textbf{chair}} \\
\toprule
{} & Chamfer & watertight & manifold \\
\midrule
ours      &              \bftab 0.7716 &     76.9\% &   97.5\% \\
ballpivot &               1.4477 &    \bftab 82.9\% & \bftab 100.0\% \\
alpha3    &               1.1966 &     53.4\% &   65.6\% \\
alpha5    &               0.9103 &     54.8\% &   60.2\% \\
\bottomrule
\end{tabular}
\end{subtable}%
\vspace{.5em}

\begin{subtable}[t]{0.45\textwidth}
\centering

\begin{tabular}{lrrr}
\multicolumn{4}{c}{\textbf{airplane}} \\
\toprule
{} & Chamfer & watertight & manifold \\
\midrule
ours      &             \bftab  0.5475 &     75.4\% &   97.3\% \\
ballpivot &               0.6295 &   \bftab  93.9\% & \bftab 100.0\% \\
alpha3    &               0.6594 &     48.0\% &   53.8\% \\
alpha5    &               1.0095 &     45.5\% &   50.4\% \\
\bottomrule
\end{tabular}
\end{subtable}\hspace{.1\textwidth}%
\begin{subtable}[t]{0.45\textwidth}
\centering

\begin{tabular}{lrrr}
\multicolumn{4}{c}{\textbf{car}} \\
\toprule
{} & Chamfer & watertight & manifold \\
\midrule
ours      &              \bftab 1.0552 &     73.3\% &   97.0\% \\
ballpivot &               1.5956 &    \bftab 78.7\% & \bftab 100.0\% \\
alpha3    &               1.3668 &     52.8\% &   64.5\% \\
alpha5    &               1.3501 &     49.8\% &   55.1\% \\
\bottomrule
\end{tabular}
\end{subtable}%
\vspace{.5em}

\begin{subtable}{0.45\textwidth}
\centering

\begin{tabular}{lrrr}
\multicolumn{4}{c}{\textbf{sofa}} \\
\toprule
{} & Chamfer & watertight & manifold \\
\midrule
ours      &              \bftab 0.8276 &    \bftab 77.6\% &   97.4\% \\
ballpivot &               1.9450 &     75.8\% & \bftab 100.0\% \\
alpha3    &               1.5633 &     51.0\% &   67.6\% \\
alpha5    &               1.3046 &     52.9\% &   58.5\% \\
\bottomrule
\end{tabular}
\end{subtable}\hspace{.1\textwidth}%
\begin{subtable}{0.45\textwidth}
\centering

\begin{tabular}{lrrr}
\multicolumn{4}{c}{\textbf{rifle}} \\
\toprule
{} & Chamfer & watertight & manifold \\
\midrule
ours      &             \bftab  0.4703 &     75.7\% &   97.3\% \\
ballpivot &               0.5567 &    \bftab 98.5\% & \bftab 100.0\% \\
alpha3    &               0.5510 &     49.2\% &   54.9\% \\
alpha5    &               0.7479 &     44.2\% &   49.6\% \\
\bottomrule
\end{tabular}
\end{subtable}%
\vspace{.5em}

\begin{subtable}{0.45\textwidth}
\centering

\begin{tabular}{lrrr}
\multicolumn{4}{c}{\textbf{lamp}} \\
\toprule
{} & Chamfer & watertight & manifold \\
\midrule
ours      &              \bftab 0.5009 &     79.9\% &   97.9\% \\
ballpivot &               0.8036 &    \bftab 92.0\% & \bftab 100.0\% \\
alpha3    &               0.6486 &     43.8\% &   52.0\% \\
alpha5    &               0.7780 &     41.7\% &   47.1\% \\
\bottomrule
\end{tabular}
\end{subtable}\hspace{.1\textwidth}%
\begin{subtable}{0.45\textwidth}
\centering

\begin{tabular}{lrrr}
\multicolumn{4}{c}{\textbf{watercraft}} \\
\toprule
{} & Chamfer & watertight & manifold \\
\midrule
ours      &              \bftab 0.5996 &     76.6\% &   97.4\% \\
ballpivot &               0.7777 &    \bftab 91.5\% & \bftab 100.0\% \\
alpha3    &               0.7643 &     49.3\% &   56.3\% \\
alpha5    &               1.0656 &     43.0\% &   48.3\% \\
\bottomrule
\end{tabular}
\end{subtable}%
\vspace{.5em}

\begin{subtable}{0.45\textwidth}
\centering

\begin{tabular}{lrrr}
\multicolumn{4}{c}{\textbf{bench}} \\
\toprule
{} & Chamfer & watertight & manifold \\
\midrule
ours      &             \bftab  0.8119 &     74.2\% &   97.2\% \\
ballpivot &               1.0239 &    \bftab 87.6\% & \bftab 100.0\% \\
alpha3    &               0.8777 &     51.7\% &   59.7\% \\
alpha5    &               0.8638 &     48.9\% &   53.9\% \\
\bottomrule
\end{tabular}
\end{subtable}\hspace{.1\textwidth}%
\begin{subtable}{0.45\textwidth}
\centering

\begin{tabular}{lrrr}
\multicolumn{4}{c}{\textbf{speaker}} \\
\toprule
{} & Chamfer & watertight & manifold \\
\midrule
ours      &              \bftab 0.7806 &    \bftab 79.4\% &   97.6\% \\
ballpivot &               2.1313 &     74.1\% & \bftab 100.0\% \\
alpha3    &               1.5941 &     48.2\% &   67.1\% \\
alpha5    &               0.9852 &     45.5\% &   51.8\% \\
\bottomrule
\end{tabular}
\end{subtable}%

\vspace{1em}
\caption{Extended results from Table 1 of the main document, reported by class over the 10 most common classes. Chamfer values are upscaled $\times 100$ for display.}
\label{tab:results_breakdown}
\end{table*}

In Table~\ref{tab:results_breakdown} below, we give per-class results for the evaluation of our method over uniformly-sampled ShapeNetCore, as presented in Table 1 of the main document.
Our method is purely local and geometric, and does not rely on class-specific features; we always train and test on all classes simultaneously.
Evaluation statistics are shown per-class here only for the sake of analysis.

\pagebreak

\begin{figure}
\begin{subfigure}{.5\textwidth}
  \centering
  \includegraphics[width=.9\linewidth]{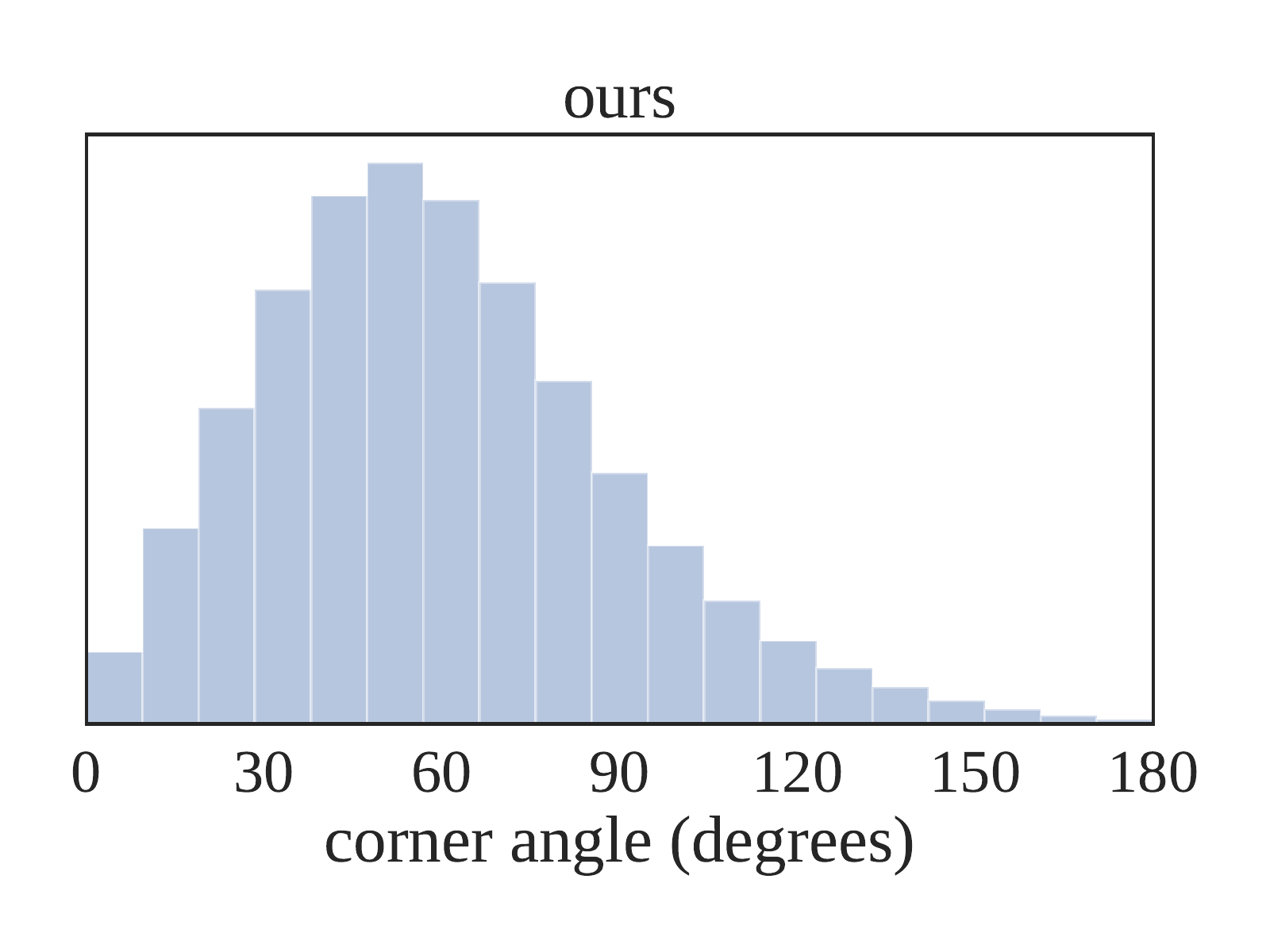}
\end{subfigure}%
\begin{subfigure}{.5\textwidth}
  \centering
  \includegraphics[width=.9\linewidth]{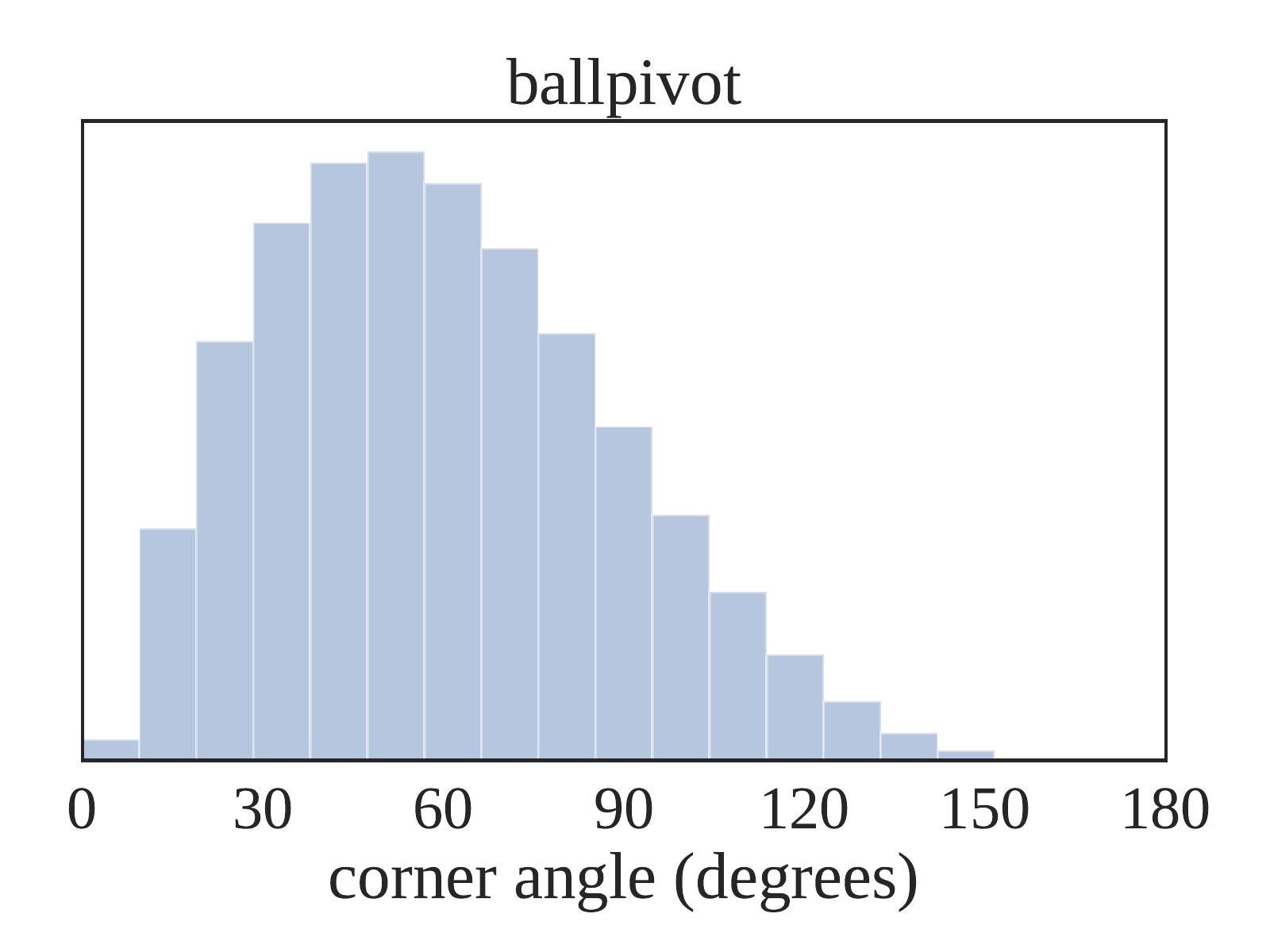}
\end{subfigure}
\begin{subfigure}{.5\textwidth}
  \centering
  \includegraphics[width=.9\linewidth]{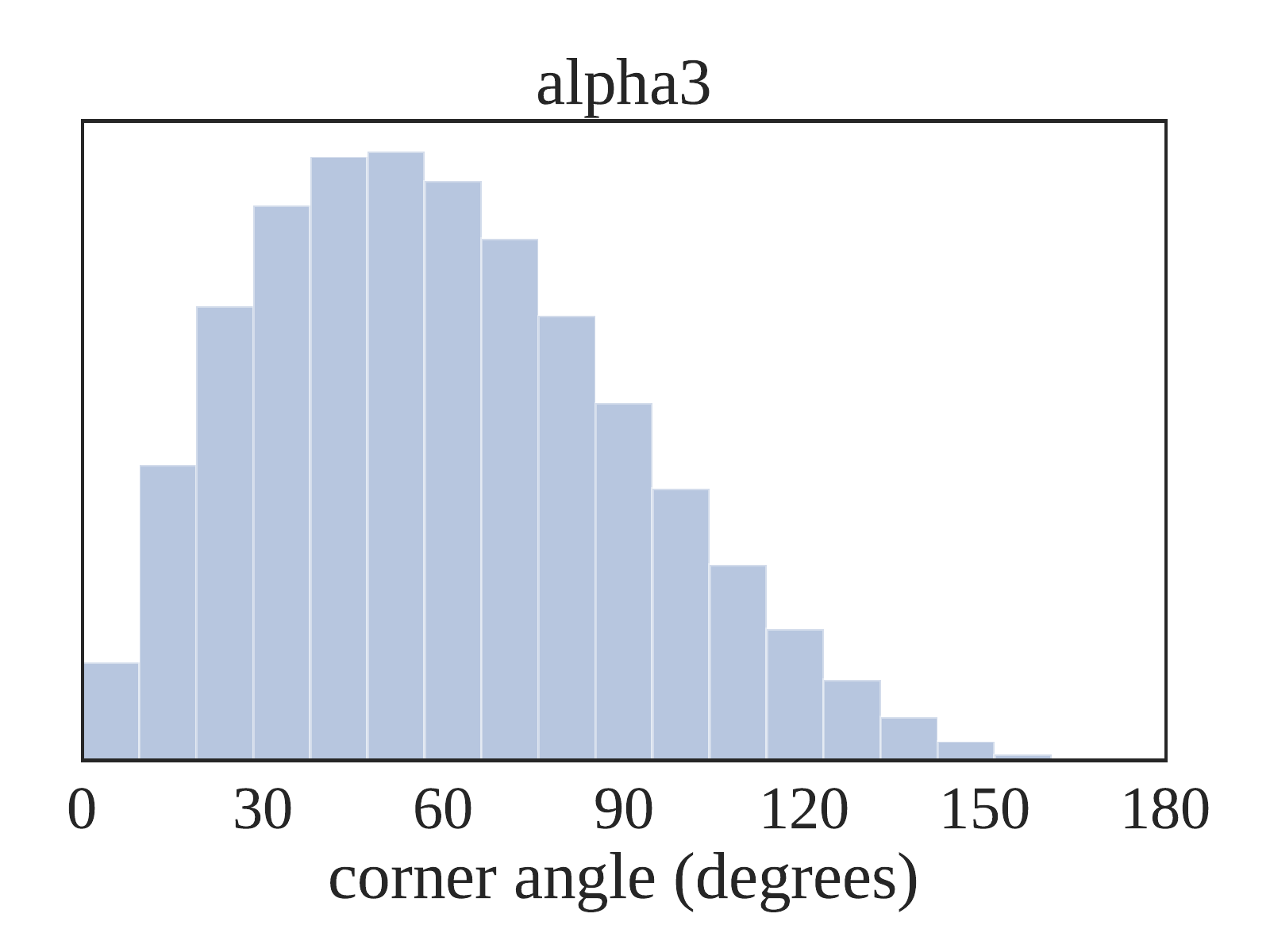}
\end{subfigure}
\begin{subfigure}{.5\textwidth}
  \centering
  \includegraphics[width=.9\linewidth]{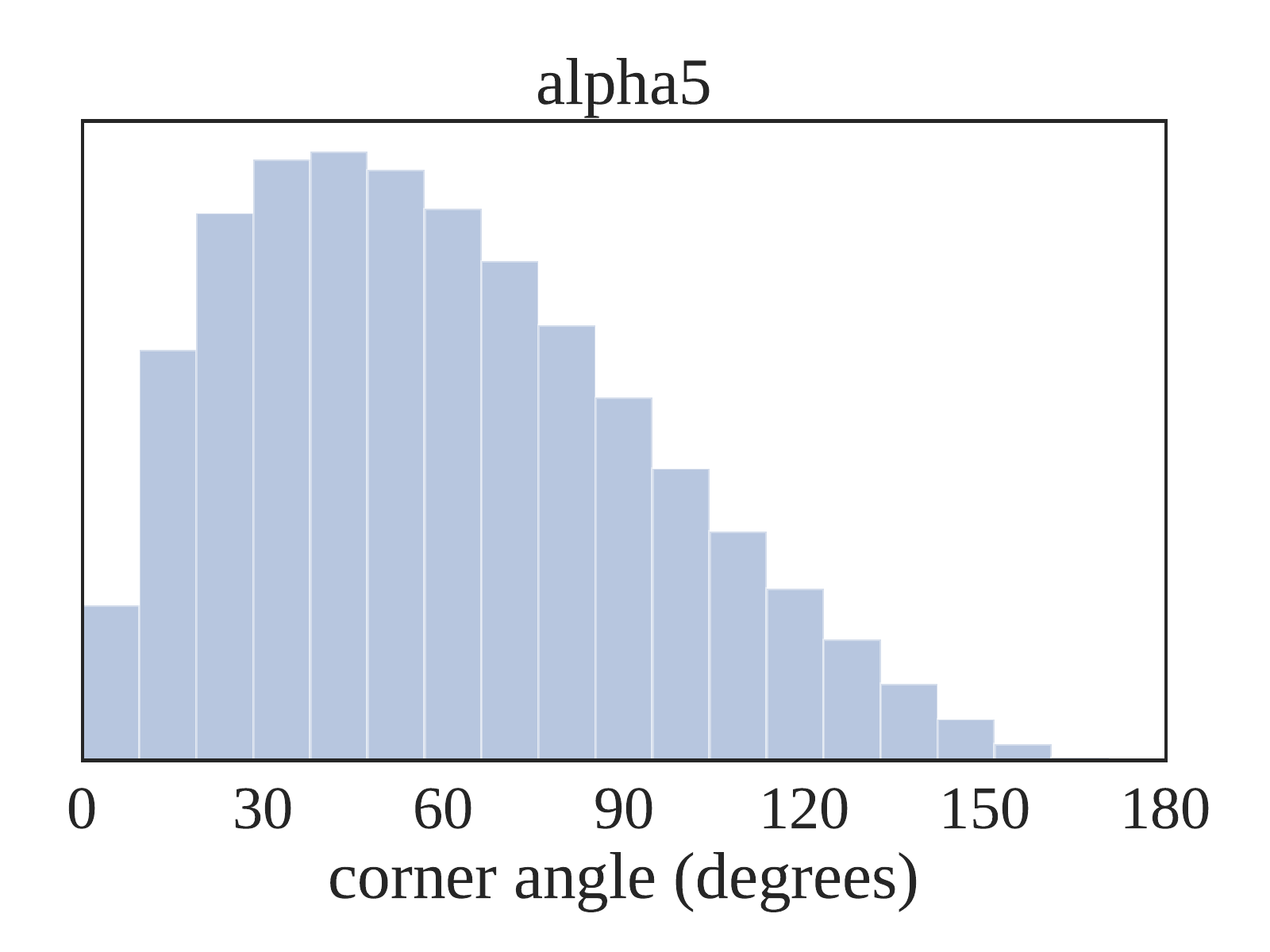}
\end{subfigure}
\caption{Histograms of corner angles in all triangles resulting from various reconstruction schemes on sampled ShapeNet, corresponding to Table 1 from the main document.}
\label{fig:angles}
\end{figure}

Triangle quality can have numerical implications for subsequent applications; extremely acute or obtuse triangles may increase approximation error or lead to poor conditioning of optimization problems.
In Figure~\ref{fig:angles}, we analyze the triangle quality in the meshes resulting from various reconstruction methods. The results are generally similar, though we observe that ball pivoting has the least prevalence of very acute triangles with angles $< 10$ degrees.
We also recall that widely-used marching cubes reconstruction likewise can yield many acute sliver triangles.

\end{document}